\documentclass[11pt,a4paper]{article}
\usepackage[hyperref]{eacl2021}
\usepackage{times}
\usepackage{latexsym}

\usepackage{multirow}
\usepackage{amsmath}
\usepackage{bm}

\usepackage{lipsum}

\usepackage{microtype}
\usepackage{textcomp}

\aclfinalcopy 

\usepackage{xcolor}

\title{
Does He Wink or Does He Nod?
A Challenging Benchmark for Evaluating Word Understanding of Language Models}

\author{Lutfi Kerem Senel {\normalfont and} Hinrich Schütze \\
  Center for Information and Language Processing (CIS), LMU Munich, Germany \\
   \texttt{lksenel@gmail.com} 
  }

\date{}

\newcounter{notecounter}
\newcommand{\enotesoff}{\long\gdef\enote##1##2{}}
\newcommand{\enoteson}{\long\gdef\enote##1##2{{
\stepcounter{notecounter}
{\large\bf
\hspace{1cm}\arabic{notecounter} $<<<$ ##1: ##2
$>>>$\hspace{1cm}}}}}
\enoteson
\enotesoff

\def\uprm#1#2{\mbox{$_#2^{\hbox{\scriptsize #1}}$}}

\hypersetup{draft}

\begin{document}
\maketitle
\begin{abstract}

Recent progress in pretraining language models on large
corpora has resulted in large performance gains on many NLP
tasks. These large models acquire linguistic knowledge
during pretraining, which helps to improve
performance on downstream tasks via fine-tuning. To assess
what kind of knowledge is acquired,
language models are commonly probed by querying them with
`fill in the blank' style cloze questions. Existing probing
datasets mainly focus on knowledge about \emph{relations between}
words and entities. We introduce WDLMPro
(Word Definition Language Model Probing) to \emph{evaluate word understanding
  directly} using dictionary definitions of
words. In our experiments, three popular pretrained
language models
struggle to match words and their  definitions. This
indicates that they understand many words poorly and that
our new probing task is a difficult challenge that could
help guide research on LMs in the future.
\end{abstract}

\section{Introduction}

Natural language processing (NLP)  has advanced drastically
in the last decade with the design of larger and more
sophisticated models, availability of larger
corpora and increasing computational
power. Pretrained word embeddings
\cite{mikolov13word2vec_b, pennington14glove} popularized
the use of distributed word representations, which became a
fundamental building block for NLP systems. 
\citet{peters18ELMO} introduced LSTM-based deep
contextual representations 
and obtained 
large performance gains 
by fine-tuning on  tasks after
unsupervised pretraining \cite{radford18fineTuning,
  howard18ULMFiT}. More recently, the attention based
transformer architecture was shown to use context more
effectively \cite{vaswani17transformers} and several
subsequent models achieved state of the art results in many
NLP tasks by combining the transformer architecture with
unsupervised pretraining and task specific fine-tuning
\cite{devlin19BERT, liu19RoBERTa}. \newcite{Radford19GPT2}
showed that language models can be applied to a variety of
tasks without task specific fine tuning. This is demonstrated on a much larger scale by \newcite{brown20GPT3}. 


Deep models improve performance. However,
what they actually learn about language and word meaning  is
still to a large extent unclear due to their uninterpretable
nature. For static word embeddings, researchers used word
similarity \cite{hill15simlex} and word analogy
\cite{gladkova16analogy} tests to shed light on what
information is captured in these dense vector spaces. For
language models, a great amount of linguistic knowledge is
stored in the model parameters \cite{peters18dissecting}.
Several studies proposed using `fill in the blank' type
cloze statements to test  knowledge learned by these models
during unsupervised pretraining. \newcite{petroni19LMasKB}
proposed the LAMA (LAnguage Model Analysis) probe to test
the factual and common sense knowledge stored in  language
models.
Similarly, \citet{Schick20rareWords} introduced WNLaMPro
(WordNet Language Model Probing)  to assess the ability of
language models to understand words based on their
frequency. In WNLaMPro, cloze style questions are generated
based on antonym, hypernym and cohyponym relations among
words extracted from WordNet.

\begin{table*}
\small
    \centering
    \begin{tabular}{l|l}
    \hline
    \textbf{synset} & \textbf{definition} \\ \hline
    \emph{a\_cappella\_singing.n.01} & \emph{singing without instrumental accompaniment} \\
     caroling.n.01 & singing joyful religious songs (especially at Christmas) \\
     crooning.n.01 & singing in a soft low tone \\
     singalong.n.01 & informal group singing of popular songs \\
     bel\_canto.n.01 & a style of operatic singing \\ \hline
   
    \end{tabular}
    \caption{Five candidates from ${\cal G}(t)$ for $t$=  \emph{a\_cappella\_singing.n.01} and their definitions} 
    \label{tab:dataset_samples}
\end{table*}

\begin{table}
    \centering
    \begin{tabular}{l|cc}
    \hline
         & \textbf{Noun} & \textbf{Verb} \\ \hline
         \textbf{\# of Synset Groups} & 51260 & 8487 \\
         \textbf{Average \# of Candidates} & 50.2 & 47.7 \\
         \textbf{min / max \# of Candidates} & 5 / 404 & 5 / 593 \\ \hline
    \end{tabular}
    \caption{WDLMPro statistics}
    \label{tab:dataset_stats}
\end{table}

The existing probing datasets mainly focus on investigating
the knowledge about \emph{relations between} words or
entities. However, a more direct way of testing whether a
language model understands the meaning of a word is to use
its dictionary definition. If a pretrained language model
truly understands the meaning of a word, then it should be
able to match it with its dictionary definition.  Based
on this motivation, we introduce the \textit{Word
  Definition Language Model Probing} (WDLMPro) dataset;\footnote{WDLAMPro and evaluation scripts are available at \href{https://www.cis.lmu.de/definition_benchmark/WDLAMPro.zip}{https://www.cis.lmu.de/definition\_benchmark/WDLAMPro.zip}} it
is a challenging benchmark for testing NLP models for their
ability to understand words.  WDLMPro is essentially a set
of thousands of synset groups; each synset group consists of
a target word (with its definition) and its taxonomic
sisters (with their definitions).
Using taxonomic sisters, rather than random word groups, 
makes the task more challenging for statistical models 
that are based on the distributional hypothesis since 
these words  have similar distributional characteristics
\cite{lenci08distributional}. We evaluate two masked
language models, BERT and RoBERTa, and the auto-regressive
model GPT-2 on WDLMPro using two different probing
tests: (i) match definition to word (D2W)
  (ii) match word to definition (W2D). We also provide a
  baseline using static fastText embeddings
  \cite{mikolov18fastText}.  We find that all three
language models perform clearly
better than the baseline. Nevertheless, they have great
difficulty matching words and their definitions, implying a
poor understanding of word meaning.  This is an important
result that could help guide research on LMs in the future.

\section{WDLMPro}
In this section,
we introduce  WDLMPro (Word Definition Language Model
Probing), a dataset to test how well NLP models can match
nouns and verbs with  their
definitions.
We view this as a test of how well the models understand lexical meaning.

\subsection{Dataset}
WordNet \cite{miller95wordnet} is the basis for
constructing WDLMPro.  A WordNet \textit{synset} contains a
set of synonyms along with a short definition of
the synset.  Different senses of polysemous words are
represented in different synsets providing
disambiguation. WordNet connects synsets with each other via
semantic relations.

Based on a \emph{target synset}
$t$ and the semantic relation hyponymy $<$, we construct
a \emph{synset group} $\cal G$ for the target as follows.
\[
  {\cal G}(t) = \{ x | \exists y: t<y \wedge x<y \}
  \]
that is, {\cal G} contains all synsets that are ``sister
hyponyms'' to $t$ with respect to a hypernym of $t$.
${\cal G}(t)$, along with the definitions of the synsets in
${\cal G}(t)$,
will be used to
set up the WDLMPro tasks that require matching of words and definitions.
We discard groups ${\cal G}(t)$ that
have a size of less than 5.

In this study, we focus on nouns and verbs, i.e., we create
synset groups ${\cal G}$ for the nouns and verbs in WordNet.
Table
\ref{tab:dataset_samples} displays five members from ${\cal
  G} (t)$ 
and their definitions
for the target
\emph{a\_cappella\_singing.n.01} (see appx. for the target \emph{beckon.v.01}.)
Table \ref{tab:dataset_stats} shows statistics of the dataset.

\subsection{Probing Tests}

We define two probing tests that are converses of each other:
\begin{itemize}
  \item \textbf{Match definition to word (D2W).} Given a
    definition and a set of words, the task is to find the
    word that the definition defines.
  \item \textbf{Match word to definition (W2D).} Given a
    word and a set of definitions, the task is to find the
    definition that defines the word.
    \end{itemize}
Each synset group ${\cal G}(t)$ gives rise to one instance
of D2W by providing the definition of $t$, and all words in
${\cal G}(t)$. The word from ${\cal G}(t)$ that matches the
definition has then to be identified. (Note that $t$ is a
member of ${\cal G}(t)$.)
Similarly, each synset group ${\cal G}(t)$ gives rise to one instance
of W2D by providing $t$ and the definitions of all words in
${\cal G}(t)$. The
correct definition  of $t$ has then to be identified among
all definition  candidates. Note that WordNet definitions by
construction do not contain the word to be defined. So there
are no instances where the two tasks are trivial.

\subsubsection{Application to language models}
In principle, any NLP model can be tested on D2W and W2D.
In this paper, we are particularly
interested in testing language models. To this end, we 
convert the data to a format that is suitable for language
models, i.e., to cloze-style questions as
shown in Table
\ref{tab:patterns}. The basic quantity that allows us to
assess the compatibility of a word $t$ and a definition is the
probability of
$t$ being generated for 
``\underline{\hspace{3mm}}'' when the definition is substituted for $<$DEF$>$.

More precisely, we compute the probability that the string
representation of $t$ is being generated.
We will denote the string representation of synset $t$ by
$\bm{t}$. 
We obtain the string representation by removing
the word type and sense information from the name of the
synset and replacing underscores with white space. For
example, synset \emph{warm\_up.v.04}
is represented by
the string ``warm
up''.

Table \ref{tab:patterns} shows that we define different templates for masked and autoregressive language models.
For the masked language models, we average the prediction scores across patterns before ranking the candidates.

\begin{table}
\centering
\begin{tabular}{ll}
\multicolumn{2}{c}{\textbf{Masked Language Model (MLM)}} \\\hline
\multirow{3}{*}{\textbf{Noun}} & \underline{\hspace{3mm}} is \texttt{<DEF>} \\
 & \underline{\hspace{3mm}} means \texttt{<DEF>}  \\
& \underline{\hspace{3mm}} is defined as \texttt{<DEF>} \\\hline
 \multirow{2}{*}{\textbf{Verb}} & definition of \underline{\hspace{3mm}} is to \texttt{<DEF>} \\
 & to \texttt{<DEF>} is the definition of \underline{\hspace{3mm}}  \\ \hline\hline
 \multicolumn{2}{c}{\textbf{Autoregressive Language Model (ALM)}}\\ \hline
\multirow{1}{*}{\textbf{Noun}} & \texttt{<DEF>} is the definition of \underline{\hspace{3mm}} \\
  \hline 
 \multirow{1}{*}{\textbf{Verb}} &   to \texttt{<DEF>} is the definition of \underline{\hspace{3mm}} \\
\end{tabular}
\caption{Patterns used for querying language models for
  nouns and verbs.
\texttt{<DEF>} refers to the definition,
\underline{\hspace{3mm}} is the mask or missing word that
the language model has to predict.}
\label{tab:patterns}
\end{table}

\subsection{Baselines}

  

For a masked language model (MLM)
$M$, the probability of a candidate $c \in {\cal G}(t)$ on W2D  is calculated as:
\begin{equation*}
  P\uprm{W2D}{M} (c|t) = \prod_{i=1}^{|\bm{t}|}P(\bm{t}^i|Q(c,|\bm{t}|))
\end{equation*}
where $\bm{t} = [\bm{t}^1, \bm{t}^2,...,\bm{t}^{|\bm{t}|}]$
is the tokenization produced by  $M$. $Q(c,|\bm{t}|)$ is the
input query
created from one of the patterns (Table \ref{tab:patterns})
with \underline{\hspace{3mm}} replaced with
$|\bm{t}|$ consecutive mask tokens. For an autoregressive
language model (ALM) $A$, we
decompose $P(\bm{t}^i|Q(c),\bm{t})$ in the standard way:
\begin{equation*}
    P\uprm{W2D}{A} = \prod_{i=1}^{|\bm{t}|}P(\bm{t}^i|Q(c),\bm{t}^{1},...,\bm{t}^{i-1})
\end{equation*}

For D2W, we need to compare, given a definition, the probabilities of
different candidate words that are generally of different lengths.
To ensure a fair comparison,
we follow
\newcite{Xiong20Encyclopedia}. For MLMs, we
match the number of mask tokens in an input query to the
token count of each candidate. The final score is the average log-probability of the masked tokens:
\begin{equation*}
    P\uprm{D2W}{M}(c|t) =
    \frac{1}{|\bm{c}|}\sum_{i=1}^{|\bm{c}|}\log P(\bm{c}^i |
    Q(t,|\bm{c}|))
\end{equation*}
For ALMs, we use the probability of
the first token:
\begin{equation*}
    P\uprm{D2W}{A}(c|t) =
     P(\bm{c}^1 |
    Q(t))
\end{equation*}
Considering further tokens does not make sense since they
are often easily predictable from the first token.


We apply our probing test to two different pretrained
MLMs (BERT and
RoBERTa) and one ALM
(GPT-2). 
To investigate the effect of model
size on the performance, we experiment with both
base and large versions of BERT and RoBERTa
along with all four sizes of GPT-2 (small, medium,
large, xl). 
For RoBERTa, we capitalize the first letter of the candidate
noun since pretrained RoBERTa models are case sensitive
and expect a capital letter at the beginning of a
sentence.\footnote{Not using capitalization resulted in poor performance for single token target words for D2W.}

In addition to the deep contextual language
models, we also provide  fastText static 
word embeddings\footnote{We use the crawl-300d-2M-subword
model from
https://fasttext.cc/docs/en/english-vectors.html}
\cite{mikolov18fastText} as a baseline.\footnote{A
reviewer suggests that it would also be interesting to
investigate the performance of supervised approaches,
e.g., ranking models. Our main
focus here is the lexical knowledge acquired in
pretraining, so we leave this for future work.} For fastText embeddings, we
tokenize the candidates and their definitions using the
NLTK tokenizer and represent them with their average
vector. We rank candidates based on their cosine
similarity to the target embedding.

\subsection{Measures}
We use two measures: precision at 1 (P@1) and a rank score
(RS), both based on a ranked results list, either of words
or of definitions. P@1 is the percentage of top-ranked items
that is correct.
We define RS as follows:  
\begin{equation*}
    \text{RS}(L,k) = \frac{L-k}{L-1}
\end{equation*}
where $L=|{\cal G}(t)|$ is the number of candidates  and
$k$ is the rank of the correct item, $1 \leq k \leq L$.
Table \ref{tab:dataset_stats} shows that the size of ${\cal G}(t)$
is highly variable;
in contrast to P@1, RS is less affected by
this and the random baseline (cf.\ Tables
\ref{tab:results_W2D}
and \ref{tab:results_D2W}) is always 0.5.  

\section{Results}

Tables \ref{tab:results_W2D} and 
\ref{tab:results_D2W} present
W2D and D2W
results for
BERT, RoBERTa and GPT-2 along with fastText and random baselines. Language models perform clearly better than both baselines.
Larger models perform generally better than smaller
ones and RoBERTa consistently outperforms BERT.
This might be an indication for the correlation between performance on WDLAMPro and downstream performance. However, further investigation is necessary to show the correlation more clearly.
For W2D, best performance is achieved by GPT-2$_{xl}$  for nouns (47.3 P@1, 0.81 RS) and by RoBERTa large for  verbs (50.8 P@1, 0.84 RS). 
Performance on D2W is much lower than for W2D for all models. 
For nouns, RoBERTa large and GPT-2$_{xl}$ perform similarly (28.8 and 29.8 P@1, 0.70 and 0.73 RS) while RoBERTa large achieves the best results for verbs (38.6 P@1, 0.80 RS).
Poor performance on D2W compared to W2D might be due to language models' ability to distinguish different definitions better than individual words since definitions are more informative  than individual words. 
Overall GPT-2 models perform better than
masked language models (with the exception of Roberta
large for verbs), despite using a single pattern as
opposed to the multiple patterns used by masked language
models. This might indicate that the ALM objective is
better at learning word meaning than the MLM objective.

\begin{table}
    \centering
    \begin{tabular}{l|rrrr}
        \hline
         \multirow{2}{*}{\textbf{Model}} & \multicolumn{2}{c}{\textbf{Noun}} & \multicolumn{2}{c}{\textbf{Verb}} \\
         & \multicolumn{1}{c}{P@1} & \multicolumn{1}{c}{RS} & \multicolumn{1}{c}{P@1} & \multicolumn{1}{c}{RS} \\ \hline
     Bert$_{b}$ & 35.2 & 0.74 & 35.3 & 0.74 \\
     Bert$_{l}$ & 35.1 & 0.73 & 33.6 & 0.73 \\
     Roberta$_{b}$ & 37.1 & 0.75 & 42.7 & 0.79 \\
     Roberta$_{l}$ & 42.1 & 0.78 & 50.8 & 0.84 \\ \hline
     GPT-2$_{s}$ & 38.7 & 0.76 & 45.0 & 0.80 \\
     GPT-2$_{m}$ & 41.8 & 0.77 & 43.6 & 0.80 \\
     GPT-2$_{l}$ & 45.7 & 0.80 & 48.4 & 0.83 \\
     GPT-2$_{xl}$ & 47.3 & 0.81 & 48.6 & 0.83 \\
     \hline 
     fastText & 22.5 & 0.66 & 29.1 & 0.69 \\ \hline 
     Random & 7.6 & 0.50 & 7.8 & 0.50 \\\hline
     
    \end{tabular}
    \caption{P@1 and rank score (RS) on W2D}
    \label{tab:results_W2D}
\end{table}

To investigate the effect of frequency,
we stratify words into \textit{rare} (fewer than 10 occurrences), \textit{medium} (10 to 99 occurrences) and \textit{frequent} (100 or more occurrences), based on occurrences in WWC\footnote{Targets that have more than 3 tokens (based on NLTK tokenization) are taken as rare without counting.} (Westbury Wikipedia Corpus, \newcite{WWC}), where we use WWC frequency as a substitute for the models' training corpora. 
We focus on nouns since most verbs in our dataset are relatively frequent. 
Table \ref{tab:freq_results_W2D} shows that, for W2D, all models have a poor understanding of the meaning of rare and medium words. (See appx. for D2W results.) 
Even for frequent words, P@1 is never above 55.

\begin{table}
    \centering
    \begin{tabular}{l|rrrr}
        \hline
         \multirow{2}{*}{\textbf{Model}} & \multicolumn{2}{c}{\textbf{Noun}} & \multicolumn{2}{c}{\textbf{Verb}} \\
         & \multicolumn{1}{c}{P@1} & \multicolumn{1}{c}{RS} & \multicolumn{1}{c}{P@1} & \multicolumn{1}{c}{RS} \\ \hline
     Bert$_{b}$ & 23.7 & 0.65 & 19.3 & 0.65 \\
     Bert$_{l}$ & 25.4 & 0.65 & 19.3 & 0.65 \\
     Roberta$_{b}$ & 25.7 & 0.67 & 32.6 & 0.74 \\
     Roberta$_{l}$ & 28.8 & 0.70 & 38.6 & 0.80 \\ \hline
     GPT-2$_{s}$ & 23.2 & 0.68 & 29.2 & 0.71 \\
     GPT-2$_{m}$ & 25.3 & 0.70 & 27.8 & 0.72 \\
     GPT-2$_{l}$ & 28.4 & 0.72 & 31.5 & 0.74 \\
     GPT-2$_{xl}$ & 29.8 & 0.73 & 32.8 & 0.76 \\ \hline 
     fastText & 16.5 & 0.63 & 20.3 & 0.69 \\ \hline 
     Random & 7.6 & 0.50 & 8.0 & 0.50 \\\hline
     
    \end{tabular}
    \caption{P@1 and rank score (RS) on D2W}
    \label{tab:results_D2W}
\end{table}

We additionally break down the results
based on the depth of the synsets in the WordNet
hierarchy. Specifically, we investigate the performance of
the GPT-2$_{xl}$ model on W2D for WordNet nouns, where we
take the depth of a synset group as the length of the
shortest path from the target synset to the root synset
(i.e., \textit{entity.n.01}). Table \ref{tab:WN_depth}
shows that
performance drops steadily as we go deeper in the
hierarchy. Lower levels of the WordNet hierarchy contain
many scientific terms and names of (sub)species such as types
of cattle (e.g., \textit{cattalo},
\textit{hereford}, \textit{galloway}). These results suggest that even very large LMs
lack the knowledge necessary to distinguish these terms.

\textbf{Analysis.}
The correct definition of the medium frequency verb `beckon' is  `signal with the hands or nod'. GPT-2$_{xl}$ predicts  `signal by winking'.
The correct definition of the frequent noun `roleplaying' is `acting a particular role (as in psychotherapy)' GPT-2$_{xl}$ predicts `acting the part of a character on stage'.
So GPT-2$_{xl}$ understands that beckoning is signaling and that roleplaying is acting, but it has not learned to distinguish between different types of signaling and acting.
This points to an important future goal for LMs: they should be developed to gain an understanding of words that goes beyond the current superficial state of the art.

\begin{table}
    \centering
    \begin{tabular}{l|rrrr}
        \hline
         \textbf{Depth} & \textbf{\# synsets} & \textbf{\# cand.} & \textbf{RS} & \textbf{P@1} \\ \hline
     3--5 & 2106 & 110 & 0.94 & 62.9 \\
     6--8 & 25,232 & 53 & 0.83 & 49.0 \\
     9--11 & 18,521 & 45 & 0.81 & 46.6 \\
     12--14 & 4473 & 19 & 0.74 & 37.4 \\ 
     15--19 & 928 & 13 & 0.67 & 31.5 \\ \hline
     
    \end{tabular}
    \caption{RS and P@1 results for GPT-2$_{xl}$  on W2D for
      nouns from different  depths of the WordNet hierarchy. \# of candidates, RS and P@1 are given as the average across all synsets within the given depth range. }
    \label{tab:WN_depth}
\end{table}

\begin{table}
    \centering
    \begin{tabular}{l|rrrr}
    \hline
         \textbf{Model} & \multicolumn{1}{c}{\textbf{rare}} & \multicolumn{1}{c}{\textbf{medium}} & \multicolumn{1}{c}{\textbf{frequent}} & \multicolumn{1}{c}{\textbf{all}} \\ \hline
     Bert$_{b}$ & 26.0 & 31.1 & 40.7 & 35.2 \\
     Bert$_{l}$ & 23.6 & 29.8 & 42.0 & 35.1 \\
     Roberta$_{b}$ & 30.8 & 34.7 & 40.7 & 37.1 \\
     Roberta$_{l}$ & 33.2 & 38.7 & 47.2 & 42.1 \\ \hline
     GPT-2$_{s}$ & 32.9 & 35.2 & 42.6 & 38.7 \\
     GPT-2$_{m}$ & 34.4 & 37.4 & 46.7 & 41.8 \\
     GPT-2$_{l}$ & 37.0 & 41.4 & 51.1 & 45.7 \\
     GPT-2$_{xl}$ & 37.7 & 42.7 & 53.3 & 47.3 \\ \hline
     Random & 6.6 & 7.0 & 8.2 & 7.6 \\ \hline 
     
    \end{tabular}
    \caption{P@1 scores on W2D for nouns of different
      frequency ranges}
    \label{tab:freq_results_W2D}
\end{table}

\textbf{Human performance on WDLAMPro.}  It is beyond the
scope of this paper to evaluate human performance on the
entirety of WDLAMPro. However, we provide a comparison with
human performance on a small subset to provide
an intuition about the difficulty of the task. For each of
the two tasks, 20 synset groups that have a maximum of 10
candidates are randomly sampled from WDLAMPro. Then two
native English speakers are asked to rank the
candidates. Table \ref{tab:human_eval} displays the average
performance of the human participants and the language
models on this subset. For both tasks, performance of the
best model is comparable to the average human performance.

Human performance is the upper bound for many NLP
tasks. We believe that this is not the case for WDLAMPro:
arguably, we should aim for models with an excellent
understanding of the meanings of words even if it is better
than average human understanding. Knowledge based tasks are
an analogous case: we should strive for models that know as
many facts as possible even if that performance is above
average human performance.

\begin{table}
    \centering
    \begin{tabular}{l|rrrr}
        \hline
         \multirow{2}{*}{\textbf{Model}} & \multicolumn{2}{c}{\textbf{W2D}} & \multicolumn{2}{c}{\textbf{D2W}} \\
         & \multicolumn{1}{c}{P@1} & \multicolumn{1}{c}{RS} & \multicolumn{1}{c}{P@1} & \multicolumn{1}{c}{RS} \\ \hline
     Bert$_{b}$ & 60.0 & 0.84 & 35.0 & 0.64 \\
     Bert$_{l}$ & 65.0 & 0.74 & 35.0 & 0.69 \\
     Roberta$_{b}$ & 50.0 & 0.78 & 60.0 & 0.81 \\
     Roberta$_{l}$ & 55.0 & 0.80 & 45.0 & 0.69 \\ \hline
     GPT-2$_{s}$ & 35.0 & 0.69 & 45.0 & 0.71 \\
     GPT-2$_{m}$ & 50.0 & 0.80 & 50.0 & 0.73 \\
     GPT-2$_{l}$ & 60.0 & 0.84 & 45.0 & 0.75 \\
     GPT-2$_{xl}$ & 50.0 & 0.76 & 45.0 & 0.79 \\ \hline 
     Human & 62.5 & 0.88 & 57.5 & 0.77 \\ \hline 
     
    \end{tabular}
    \caption{LM and human performance on 20 random samples of WDLAMPro. }
    \label{tab:human_eval}
\end{table}

\section{Conclusion}
We introduced
WDLMPro,
a
probing test that helps  analyze  how well a model
understands word meaning. WDLMPro is complementary to existing
probing tests that are about
\emph{relations} between words or entities.
We evaluated three popular pretrained language
models on the W2D (word to definition) and D2W (definition
to word) tasks. Our findings show
that, despite their remarkable performance on many
downstream tasks, these models struggle to match a word and
its true definition, suggesting an insufficient
understanding of word meaning.
Relatively poor performance of these
powerful models on WDLMPro can be seen as evidence
for the limitations of purely distributional systems and
the need for incorporating external knowledge.
WDLMPro provides an important
evaluation benchmark, encouraging design and training of
models with precise word understanding.

\textbf{Acknowledgements.} We thank Denis Peskov and Sander Schulhoff for
helping out with the human evaluation and
the anonymous reviewers for their insightful comments and suggestions.
This work was funded by the European Research Council (ERC \#740516).

\newpage
\appendix
\section{Appendix}
\label{sec:appendix}

\begin{table}[h]
    \centering
    \begin{tabular}{l|l}
    \hline
    \textbf{synset} & \textbf{definition} \\ \hline
     \emph{beckon.v.01} & \emph{signal with the hands or nod} \\
     applaud.v.01 & clap one's hands or shout after performances to indicate approval \\
     bow.v.01 & bend one's knee or body, or lower one's head \\
     shrug.v.01 & raise one's shoulders to indicate indifference or resignation \\
     exsert.v.01 & thrust or extend out \\
     wink.v.01 & signal by winking \\
     nod.v.01 & express or signify by nodding \\\hline
   
    \end{tabular}
    \caption{Seven candidates of ${\cal G}(t)$ for $t$=  \emph{beckon.v.01} and their definitions} 
    \label{tab:dataset_samples2}
\end{table}

\begin{table}[h]
    \centering
    \begin{tabular}{l|rrrr}
    \hline
        \textbf{Model} & \multicolumn{1}{c}{\textbf{rare}} & \multicolumn{1}{c}{\textbf{medium}} & \multicolumn{1}{c}{\textbf{frequent}} & \multicolumn{1}{c}{\textbf{all}} \\ \hline
     Bert$_{b}$ & 14.7 & 20.6 & 28.7 & 23.7 \\
     Bert$_{l}$ & 12.0 & 20.1 & 33.1 & 25.4\\
     Roberta$_{b}$ & 17.7 & 24.2 & 29.5 & 25.7 \\
     Roberta$_{l}$ & 17.9 & 25.8 & 34.5 & 28.8 \\ \hline
     GPT-2$_{s}$ & 17.3 & 20.7 & 26.7 & 23.2 \\
     GPT-2$_{m}$ & 17.0 & 21.1 & 30.6 & 25.3 \\
     GPT-2$_{l}$ & 19.2 & 24.3 & 33.9 & 28.4 \\
     GPT-2$_{xl}$ & 19.3 & 24.8 & 36.3 & 29.8 \\ \hline
     Random & 6.7 & 7.1 & 8.3 & 7.6 \\ \hline 
     
    \end{tabular}
    \caption{P@1 scores on D2W for nouns based on target word frequency.}
    \label{tab:freq_results_D2W}
\end{table}

\end{document}